# Discovering Cyclic Causal Models
# by Independent Components Analysis

**Gustavo Lacerda**
Machine Learning Department
School of Computer Science
Carnegie Mellon University
Pittsburgh, PA 15213

**Peter Spirtes**
**Joseph Ramsey**
Department of Philosophy
Carnegie Mellon University
Pittsburgh, PA 15213

**Patrik O. Hoyer**
Dept. of Computer Science
University of Helsinki
Helsinki, Finland

## Abstract

We generalize Shimizu et al's (2006) ICA-based approach for discovering linear non-Gaussian acyclic (LiNGAM) Structural Equation Models (SEMs) from causally sufficient, continuous-valued observational data. By relaxing the assumption that the generating SEM's graph is acyclic, we solve the more general problem of linear non-Gaussian (LiNG) SEM discovery. LiNG discovery algorithms output the distribution equivalence class of SEMs which, in the large sample limit, represents the population distribution. We apply a LiNG discovery algorithm to simulated data. Finally, we give sufficient conditions under which only one of the SEMs in the output class is "stable".

## 1    Linear SEMs

Linear structural equation models (SEMs) are statistical causal models widely used in the natural and social sciences (including econometrics, political science, sociology, and biology) [1].

The variables in a linear SEM can be divided into two sets, the error terms (typically unobserved), and the substantive variables. For each substantive variable $\mathbf{x}_i$, there is a linear equation with $\mathbf{x}_i$ on the left-hand-side, and the direct causes of $\mathbf{x}_i$ plus the corresponding error term on the right-hand-side.

Each SEM with jointly independent error terms can be associated with a directed graph (abbreviated as DG) that represents the causal structure of the model and the form of the linear equations. The vertices of the graph are the substantive variables, and there is a directed edge from $\mathbf{x}_i$ to $\mathbf{x}_j$ just when the coefficient of $\mathbf{x}_i$ in the structural equation for $\mathbf{x}_j$ is non-zero. [1]

[1]Traditionally, SEMs with acyclic graphs are called "re-

### 1.1    The model, with an illustration

Let $\mathbf{x}$ be the random vector of substantive variables, $\mathbf{e}$ be the vector of error terms, and $B$ be the matrix of linear coefficients for the substantive variables. Then the following equation describes the linear SEM model:

$$\mathbf{x} = B\mathbf{x} + \mathbf{e} \qquad (1)$$

For example, consider the model defined by:

$$\begin{aligned}
\mathbf{x}_1 &= \mathbf{e}_1 \\
\mathbf{x}_2 &= 1.2\mathbf{x}_1 - 0.3\mathbf{x}_4 + \mathbf{e}_2 \\
\mathbf{x}_3 &= 2\mathbf{x}_2 + \mathbf{e}_3 \\
\mathbf{x}_4 &= -\mathbf{x}_3 + \mathbf{e}_4 \\
\mathbf{x}_5 &= 3\mathbf{x}_2 + \mathbf{e}_5
\end{aligned} \qquad (2)$$

Note that the coefficient of each variable on the left-hand-side of the equation is 1.

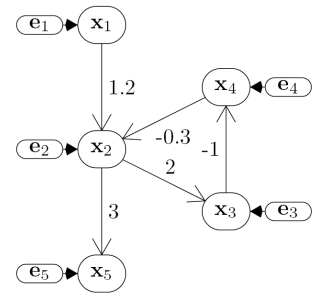

Fig. 1: Example 1

$\mathbf{x}$ can also be expressed directly as a linear combination of the error terms, as long as $I - B$ is invertible. Solving for $\mathbf{x}$ in Eq. 1 gives $\mathbf{x} = (I - B)^{-1}\mathbf{e}$. If we let $A = (I - B)^{-1}$, then $\mathbf{x} = A\mathbf{e}$. $A$ is called the reduced form matrix (in the terminology of Independent cursive", and SEMs with cyclic graphs "non-recursive"[15]. We avoid this usage, and use "acyclic" or "cyclic" instead.

Components Analysis (see Section 3.1), it is called the "mixing matrix").

The distributions over the error terms in a SEM, together with the linear equations, entail a joint distribution over the substantive variables. This joint distribution can be interpreted in terms of physical processes, as shown next.

## 1.2 Interpretating linear SEMs

These equations (contained in matrix equation (1)) can be given several different interpretations. Under one class of interpretations, they are a set of equations satisfied by a set of variables $\mathbf{x}$ in equilibrium. With some further assumptions, the $B$ matrix in the simultaneous equations (a.k.a. "equilibrium equations") also represents the coefficients in a set of dynamical equations describing a deterministic dynamical system.

Fisher [5] gave one such interpretation as follows: There is a relatively long observation period of length 1, and a much shorter "reaction" lag of length $\Delta\theta = 1/n$. The observed variable is the vector $\bar{\mathbf{x}}[t]$, defined as the average of $\mathbf{x}$ over the observation period starting at $t$:

$$\bar{\mathbf{x}}[t] \equiv \frac{1}{n}\sum_{k=1}^{n}\mathbf{x}[t + k\Delta\theta] \qquad (3)$$

Suppose that the underlying dynamical equations are:

$$\mathbf{x}[t + k\Delta\theta] = B_{dyn}\mathbf{x}[t + (k-1)\Delta\theta] + \mathbf{e} \qquad (4)$$

where $\mathbf{e}$ is constant over the observation period (but may differ for different units in the population, e.g. different observation periods).

Fisher showed that, in the limit as $\Delta\theta$ approaches 0, there is a $B_{equail} = B_{dyn}$ such that:

$$\bar{\mathbf{x}}[t] = B_{equil}\bar{\mathbf{x}}[t] + \mathbf{e} \qquad (5)$$

if and only if the modulus of each eigenvalue of $B_{dyn}$ is less than or equal to 1, and no eigenvalue is equal to 1.

The assumptions underlying this model are fairly strong, but commonly made in econometrics, and defended by Fisher [5].

A simpler, but similar interpretation with similar consequences is the one in which the observed value $\mathbf{x}$ is the state in which the dynamical system converged, rather than its average over an observation period.

### 1.2.1 Dealing with self-loops

The LiNG discovery algorithms presented in this paper (described in section 4) output a set of directed graphs that do not contain any "self-loops" (edges from a vertex to itself) [2], i.e. the $B$ matrices output by our LiNG discovery algorithms have all zeros in the diagonal. This is because it is impossible to determine the values of the diagonal entries of the $B$ matrix from equilibrium data alone.

In the underlying dynamical equations, it may be that for some index $a$, $\mathbf{x}_a[t + (k-1)\Delta\theta]$ affects $\mathbf{x}_a[t + k\Delta\theta]$ (i.e. $b_{a,a} \neq 0$). Our goal is to recover the coefficients that both represent the distribution of $\bar{\mathbf{x}}$ and correctly predict the effects of manipulations. A manipulation of a variable $\mathbf{x}_i$ to a distribution $P$ is modeled by replacing the dynamical equation for $\mathbf{x}_i$ by a new dynamical equation $\mathbf{x}_i[t + k\Delta\theta] = \mathbf{e}'_i$, where $\mathbf{e}'_i$ has distribution $P$ [10].

For these purposes, the following argument sketches why the underdetermination of the diagonal of $B_{equil}$ by the equilibrium data is not a problem, as long as $b_{a,a} \neq 1$ in the underlying dynamical equations.

The equation for $\bar{\mathbf{x}}_a$ has the form:

$$\bar{\mathbf{x}}_a = b_{a,a}\bar{\mathbf{x}}_a + \sum_{k\neq a, k=1}^{n}b_{a,k}\bar{\mathbf{x}}_k + e_a \qquad (6)$$

If $b_{a,a} \neq 1$, it is possible to rewrite this as:

$$\bar{\mathbf{x}}_a - b_{a,a}\bar{\mathbf{x}}_a = \sum_{k\neq a, k=1}^{n}b_{a,k}\bar{\mathbf{x}}_k + e_a \qquad (7)$$

$$\bar{\mathbf{x}}_a = \frac{1}{1 - b_{a,a}}\left(\sum_{k\neq a, k=1}^{n}b_{a,k}\bar{\mathbf{x}}_k + e_a\right) = \sum_{k=1}^{n}b'_{a,k}\bar{\mathbf{x}}_k + e'_a \qquad (8)$$

where $b'_{a,a} = 0$. The modified system of equations containing Equation 8 is represented by a graph that has no self-loops, and has a different underlying dynamical equation in which the coefficient for $\mathbf{x}_a[t + (k-1)\Delta\theta]$ in the equation for $\mathbf{x}_a[t + k\Delta\theta]$ is zero.

Note that in the second equation, the error term $e_a$ has been rescaled by $1/(1 - b_{a,a})$ to form a new error term $e'_a$ and when $(I - B)^{-1}$ is taken to form the reduced form coefficients, the coefficients corresponding to $e_a$ in the first set of equations will be multiplied by $(1 - b_{a,a})$, and the two changes cancel each other out.

Now, if we consider the original dynamical system and the one that results from setting the diagonal of $B$ to zero (as above), it is sometimes the case that one dynamical system satisfies the conditions for the dynamical equations to approach the simultaneous equations in the limit, while the other one does not, because the magnitude of the coefficients in the equation for $\mathbf{x}_a[t]$ are different. If both forms satisfy Fisher's conditions, then the act of manipulating any variable to a fixed distribution for all $t$ makes the two sets of dynamical equations have equivalent limiting simultaneous equations.

---

[2]Fisher argues that self-loops are not realistic, but these arguments are not entirely convincing.

### 1.2.2 Self-loops with coefficient 1

Unfortunately, the case where $b_{a,a} = 1$ cannot be handled in the same way, since $1/(1 - b_{a,a})$ is infinite. If $b_{a,a} = 1$, then there may be no equivalent form without a self-loop (or more precisely, the corresponding equations without a self-loop may require setting the variance of some error terms to zero). The case where $b_{a,a} = 1$ is a genuine problem that we do not currently have a solution for. For the purposes of this paper, we assume that no self-loops have a coefficient of 1.

As Dash has pointed out [4], there are cases where the simultaneous equations have a different graph than the underlying dynamical equations, and hence the graph that represents the simultaneous equations cannot be used to predict the effects of a manipulation of the underlying dynamical system. In [4], Dash presents two such examples. In both of them, in effect, $B_{dyn}$ has a 1 in the diagonal.

## 2 The problem and its history

### 2.1 The problem of DG causal discovery

Using the interpretations from 1.2, we can frame the problem as follows: given samples of the equilibrium distribution of a LiNG process whose observed variables form a causally sufficient set [3], find the set of SEMs that describe this distribution, under the assumption that it is non-empty.

### 2.2 Richardson's Cyclic Causal Discovery (CCD) Algorithm

While many algorithms have been suggested for discovering (equivalence classes of) directed acyclic graphs (DAGs) from data, for general linear directed graphs (DGs) only one provably correct algorithm was known (until now), namely Richardson's Cyclic Causal Discovery (CCD) algorithm.

CCD outputs a "partial ancestral graph" (PAG) that represents both a set of directed graphs that entail the same set of zero partial correlations for all values of the linear coefficients, and features common to those directed graphs (such as ancestor relations). The algorithm performs a series of statistical tests of zero partial correlations to construct the PAG. The set of zero partial correlations that is entailed by a linear SEM with uncorrelated errors depends only upon the linear coefficients, and not upon the distribution of the error terms. Under some assumptions [4], in the large sample limit, CCD outputs a PAG that represents the true graph.

There are a number of limitations to this algorithm. First, the set of DGs contained in a PAG can be large, and while they all entail the same zero partial correlations (viz., those judged to hold in the population), they need not entail the same joint distribution or even the same covariances. Hence in some cases, the set represented by the PAG will include cyclic graphs that do not fit the data well. Therefore, even assuming that the errors are all Gaussian, it is possible to reduce the size of the set of graphs output by CCD, although in practice this can be intractable. For details on the algorithm, see [11].

## 3 Shimizu et al's approach for discovering LiNGAM SEMs

The "LiNGAM algorithm"[12], which uses Independent Components Analysis (ICA), reliably discovers a unique correct LiNGAM SEM, under the following assumptions about the data: the structural equations of the generating process are linear and can be represented by an acyclic graph; the error terms have non-zero variance; the samples are independent and identically distributed; no more than one error term is Gaussian; and the error terms are jointly independent.[5]

### 3.1 Independent Components Analysis (ICA)

Independent components analysis [3, 8] is a statistical technique used for estimating the mixing matrix $A$ in equations of the form $\mathbf{x} = A\mathbf{e}$ ($\mathbf{e}$ is often called "sources" and written $\mathbf{s}$), where $\mathbf{x}$ is observed and $\mathbf{e}$ and $A$ are not.

ICA algorithms find the invertible linear transforma-tion $W = A^{-1}$ of the data $X$ that makes the error distributions corresponding to the implied samples $E$ of **e** maximally non-Gaussian (and thus, maximally independent). The matrix $A$ can be identified up to scaling and permutation as long as the observed distribution is a linear, invertible mixture of independent components, at most one of which is Gaussian [3]. There are computationally efficient algorithms for estimating $A$ [8].

---

[3]A set $V$ of variables is causally sufficient for a population if and only if in the population every common direct cause of any two or more variables in $V$ is in $V$. (For subtleties regarding this definition, see [13]).

[4]The assumptions are: the samples are independent and identically distributed, no error term has zero variance, the statistical tests for zero partial correlations are consistent, linearity of the equations, the existence of a unique reduced form, faithfulness (i.e. there are no zero partial correlations in the population that are not entailed for all values of the free parameters of the true graph), and that the error terms are uncorrelated.

[5]The error terms are typically not jointly independent if the set of variables is not causally sufficient.

### 3.2 The LiNGAM discovery algorithm

If we run an ICA algorithm on data generated by a linear SEM, the matrix $W_{ICA}$ obtained will be a row-scaled, row-permuted version of $I - B$, where $B$ is the coefficient matrix of the true model (this is a consequence of the derivation in Section 1.1). We are now left with the problem of finding the proper permutation and scale for the $W$ matrix so that it equals $I - B$.

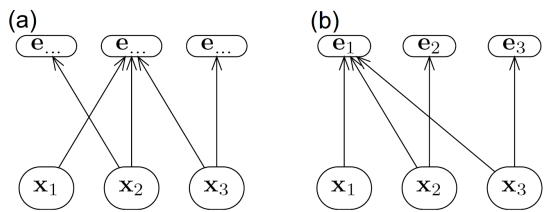

Fig. 2: After removing the edges whose coefficients are statistically indistinguishable from zero: (a) the raw $W_{ICA}$ matrix output by ICA on a SEM whose graph is $\mathbf{x}_2 \to \mathbf{x}_1 \leftarrow \mathbf{x}_3$ (b) the corresponding $\widetilde{W}$ matrix, obtained by permuting the error terms in $W_{ICA}$

Since the order of the error terms given by ICA is arbitrary, the algorithm needs to correctly match each error term $\mathbf{e}_i$ to its respective substantive variable $\mathbf{x}_i$. This means finding the correct permutation of the rows of $W_{ICA}$. We know that the row-permutation of $W_{ICA}$ corresponding to the correct model cannot have a zero in the diagonal (we call such permutations "inadmissible") because $W = I - B$, and the diagonal of $B$ is zero.

Since, by assumption, the data was generated by a DAG, there is exactly one row-permutation of $W_{ICA}$ that is admissible [12]. To visualize this, this constraint says that there is exactly one way to reorder the error terms so that every $\mathbf{e}_i$ is the target of a vertical arrow.[6]

In this example, swapping the first and second error terms is the only permutation that produces an admissible matrix, as seen in Fig. 2(b).

---

[6] Another consequence of acyclicity is that there will be no right-pointing arrows in this representation, provided that the **x**s are topologically sorted w.r.t. the DAG.

After the algorithm finds the correct permutation, it finds the correct scaling, i.e. "normalizing" $W$ by dividing each row by its diagonal element, so that the diagonal of the output matrix is all 1s (i.e. the coefficient of each error term is 1, as specified in Section 1).

Bringing it all together, the algorithm computes $B$ by using $B = I - W'$, where $W' = normalize(\widetilde{W})$, $\widetilde{W} = RowPermute(W_{ICA})$ and $W_{ICA} = ICA(X)$.

Besides the fact that it determines the direction of every causal arrow, another advantage of LiNGAM over conditional-independence-based methods [13] is that the correctness of the algorithm does not require the faithfulness assumption.

For more details on the LiNGAM approach, see [12].

## 4 Discovering LiNG SEMs

The assumptions of the family of LiNG discovery algorithms described below (abbreviated as "LiNG-D") are the same as the LiNGAM assumptions, replacing the assumption that the SEM is acyclic with the weaker assumption that the diagonal of $B_{dyn}$ contains no 1s. In this more general case, as in the acyclic case, candidate models are generated by finding all admissible matches of the error terms ($\mathbf{e}_i$'s) to the observed variables ($\mathbf{x}_i$'s). In other words, each candidate corresponds to a row-permutation of the $W_{ICA}$ matrix that has a zeroless diagonal.

As in LiNGAM, the output is the set of admissible models. In LiNGAM, this set is guaranteed to contain a single model, thanks to the acyclicity assumption. If the true model has cycles, however, more than one model will be admissible.

The remainder of this section addresses the problem of finding the admissible models, given that ICA has finite data to work with.

### 4.1 Prune and solve Constrained n-Rooks

These algorithms generate candidate models by testing which entries of $W_{ICA}$ are zero (i.e. pruning), and finding all admissible permutations based on that (i.e. solving Constrained n-Rooks, see Section 4.1.2). We call an algorithm "local" if, for each entry $w_{i,j}$ of $W_{ICA}$, it makes a decision about whether $w_{i,j}$ is zero using only $w_{i,j}$.

#### 4.1.1 Deciding which entries are zero

There are several methods for deciding which entries of $W_{ICA}$ to set to zero:

- **Thresholding**: the simplest method for estimating which entries of $W_{ICA}$ are zero is to simply choose a threshold value, and set every entry of $W_{ICA}$ smaller than the threshold (in absolute value) to zero. This method fails to account for the fact that different coefficients may have different spreads, and will miss all coefficients smaller than the threshold.

- **Test the non-zero hypothesis by bootstrap sampling**: another method for estimating which entries of $W_{ICA}$ are actually zero is to do bootstrap sampling. Bootstrap samples are created by resampling with replacement from the original data. Then ICA is run on each bootstrap sample, and each coefficient $w_{i,j}$ is calculated for each bootstrap sample. This leads to a real-valued distribution for each coefficient.[7] Then, for each one, a non-parametric quantile test is performed in order to decide whether 0 is an outlier. If it isn't, the coefficient is set to 0 (i.e. the corresponding edge is pruned.)[8]

- **Use sparse ICA**: Use an ICA algorithm that returns a sparse (i.e. pre-pruned) mixture, such as the one presented by Zhang and Chan [16]. Unlike the other methods above, this is not a local algorithm.

### 4.1.2 Constrained n-Rooks: the problem and an algorithm

Once it is decided which entries are zero, the algorithm searches for every row-permutation of $W_{ICA}$ that has a zeroless diagonal. Each such row-permutation corresponds to a placement of $n$ rooks onto the non-zero entries on an $n \times n$ chessboard such that no two rooks threaten each other. Then the rows are permuted so that all the rooks end up on the diagonal, thus ensuring that the diagonal has no zeros.

To solve this problem, we use a simple depth-first search that prunes search paths that have nowhere to place the next rook. In the worst case, every permutation is admissible, and the search must take $O(n!)$.

---

[7]One needs to be careful when doing this, since each run of ICA may return a $W_{ICA}$ in a different row-permutation. This means that we first need to row-permute each bootstrap $W_{ICA}$ to match with the original $W_{ICA}$.

[8]One could object that, instead of a quantile test, the correct procedure would be to simulate under the null hypothesis (i.e.: edge is absent) using the estimated error terms, and then compare the obtained distribution of the ICA statistics with their distribution for the bootstrap. However, this raises issues and complexities that are tangential to the current paper.

### 4.2 A non-local algorithm

Local algorithms work under the assumption that the estimates of the $w_{i,j}$ are independent of each other – which is in general false when estimating with finite samples. This motivates the use of non-local methods.

In the LiNGAM (acyclic) approach [12], a non-local algorithm is presented for finding the single best row-permutation of $W_{ICA}$, which minimizes a loss function that heavily penalizes entries in the diagonal that are close to zero (such as $x \rightarrow |1/x|$). This is written as a linear assignment problem (i.e. finding the best match between the $\mathbf{e}_i$s and $\mathbf{x}_i$s), which can be solved using the Hungarian algorithm [9] or others.

For general LiNG discovery, however, algorithms that find the best linear assignment do not suffice, since there may be multiple admissible permutations.

One idea is to use a $k$-th best assignment algorithm [2] (i.e. the $k$-th permutation with the least penalty on the diagonal), for increasing $k$. With enough data, all permutations corresponding to inadmissible models will score poorly, and there should be a clear separation between admissible and inadmissible models.

The non-local method presented above, like the thresholding method, fails to account for differences in spread among estimates of the entries of $W_{ICA}$. It would be straightforward to fix this by modifying the loss function to penalize diagonal entries for which the test fails to reject the null hypothesis (as described in the part about bootstrap sampling in Section 4.1.1), instead of penalizing them for merely being close to zero.

### 4.3 Sample run

We generated 15000 sample points using the SEM in Example 1 and error terms distributed according to a symmetric Gaussian-squared distribution[9].

Fig. 3 shows the output of the local thresholding algorithm with the cut-off set to 0.05.

For the sake of reproducibility, our code with instructions is available from: `www.phil.cmu.edu/~tetrad/cd2008.html`.

## 5 Theory

### 5.1 Notions of DG equivalence

There are a number of different senses in which the directed graphs associated with SEMs can be "equivalent" or "indistinguishable" given observational data,

---

[9]The distribution was created by sampling from the standard Gaussian(0,1) and squaring it. If the value sampled was negative, it was made negative again.

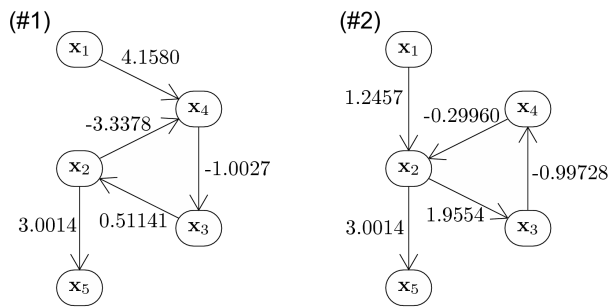

Fig. 3: The output of LiNG-D: Candidate #1 and Candidate #2

assuming linearity and no dependence between error terms:

- DGs $G_1$ and $G_2$ are *zero partial correlation equivalent* if and only if the set of zero partial correlations entailed for all values of the free parameters (non-zero linear coefficients, distribution of the error terms) of a linear SEM with DG $G_1$ is the same as the set of zero partial correlations entailed for all values of the free parameters of a linear SEM with $G_2$. For linear models, this is the same as *d-separation equivalence*. [13]

- DGs $G_1$ and $G_2$ are *covariance equivalent* if and only if for every set of parameter values for the free parameters of a linear SEM with DG $G_1$, there is a set of parameter values for the free parameters of a linear SEM with DG $G_2$ such that the two SEMs entail the same covariance matrix over the substantive variables, and vice-versa.

- DGs $G_1$ and $G_2$ are *distribution equivalent* if and only if for every set of parameter values for the free parameters of a linear SEM with DG $G_1$, there is a set of parameter values for the free parameters of a linear SEM with DG $G_2$ such that the two SEMs entail the same distribution over the substantive variables, and vice-versa. Do not confuse this with the notion of distribution-entailment equivalence between SEMs: two SEMs with fixed parameters are *distribution-entailment equivalent* iff they entail the same distribution.

It follows from well-known theorems about the Gaussian case [13], and some trivial consequences of known results about the non-Gaussian case [12], that the following relationships exist among the different senses of equivalence for acyclic graphs: If all of the error terms are assumed to be Gaussian, distribution equivalence is equivalent to covariance equivalence, which in turn is equivalent to d-separation equivalence. If not all of the error terms are assumed to be Gaussian, then distribution equivalence entails (but is not entailed by) covariance equivalence, which entails (but is not entailed by) d-separation equivalence.

So for example, given Gaussian error terms, $A \leftarrow B$ and $A \rightarrow B$ are zero partial correlation equivalent, covariance equivalent, and distribution equivalent. But given non-Gaussian error terms, $A \leftarrow B$ and $A \rightarrow B$ are zero-partial-correlation equivalent and covariance equivalent, but not distribution equivalent. So for Gaussian errors and this pair of DGs, no algorithm that relies only on observational data can reliably select a unique acyclic graph that fits the population distribution as the correct causal graph without making further assumptions; but for all (or all except one) non-Gaussian errors there will always be a unique acyclic graph that fits the population distribution.

While there are theorems about the case of cyclic graphs and Gaussian errors, we are not aware of any such theorems about cyclic graphs with non-Gaussian errors with respect to distribution equivalence. In the case of cyclic graphs with all Gaussian errors, distribution equivalence is equivalent to covariance equivalence, which entails (but is not entailed by) d-separation equivalence [14]. In the case of cyclic graphs in which at most one error term is non-Gaussian, distribution equivalence entails (but is not entailed by) covariance equivalence, which in turn entails (but is not entailed by) d-separation equivalence. However, given at most one Gaussian error term, the important difference between acyclic graphs and cyclic graphs is that no two different acyclic graphs are distribution equivalent, but there are different cyclic graphs that are distribution equivalent.

Hence, no algorithm that relies only on observational data can reliably select a unique cyclic graph that fits the data as the correct causal graph without making further assumptions. For example, the two cyclic graphs in Fig. 3 are distribution equivalent.

## 5.2 The output of LiNG-D is correct and as fine as possible

**Theorem 1** *The output of LiNG-D is a set of SEMs that comprise a distribution-entailment equivalence class.*

**Proof:** First, we show that any two SEMs in the output of LiNG-D entail the same distribution.

The weight matrix output by ICA is determined only up to scaling and row permutation. Intuitively, then, permuting the error terms does not change the mixture. Now, more formally:

Let $M_1$ and $M_2$ be candidate models output by LiNG-D. Then $W_1$ and $W_2$ are row-permutations of $W_{ICA}$: $W_1 = P_1 W_{ICA}$, $W_2 = P_2 W_{ICA}$

Likewise, for the error terms: $E_1 = P_1 E$, $E_2 = P_2 E$

Then the list of samples $X$ implied by $M_1$ is $A_1 E_1 = (W_1)^{-1} E_1 = (P_1 W_{ICA})^{-1} (P_1 E) = W_{ICA}^{-1} P_1^{-1} P_1 E = W_{ICA}^{-1} E$.

By the same argument, the list of samples $X$ implied by $M_2$ is also $W_{ICA}^{-1} E$. Therefore, any two SEM models output by LiNG-D entail the same distribution.

Now, it remains to be shown that if LiNG-D outputs one SEM that entails a distribution $P$, it outputs all SEMs that entail $P$.

Suppose that there is a SEM $S$ that represents the same distribution as some $T$, which is output by LiNG-D. Then the reduced-form coefficient matrices for $S$ and $T$, $A_S$ and $A_T$, are the same up to column-permutation and scaling. Hence, $I - B_S$ and $I - B_T$ are also the same up to scaling and row-permutation (by $I - B = A^{-1}$). By the assumption that there are no self-loops with coefficient 1, neither $I - B_T$ nor $I - B_S$ has zeros on the diagonal. Since $I - B_T$ is a scaled row-permutation of $W_{ICA}$ that has no zeros on the diagonal, so is $I - B_S$. Thus $S$ is also output by LiNG-D. □

**Theorem 2** *If the simultaneous equations are linear and can be represented by a directed graph; the error terms have non-zero variance; the samples are independently and identically distributed; no more than one error term is Gaussian; and the error terms are jointly independent, then in the large sample limit, LiNG-D outputs all SEMs that entail the population distribution.*

**Proof:** ICA gives pointwise consistent estimates of $A$ and $W$ under the assumptions listed [3]. This entails that there are pointwise consistent tests of whether an entry in the $W$ matrix is zero, and hence (by definition) in the large sample limit, the limit of both type I and type II errors of tests of zero coefficients are zero. Given the correct zeroes in the $W$ matrix, the output of the local version of the LiNG-D algorithm is correct in the sense that the simultaneous equation describes the population distribution. □

In general, each candidate model $B' = I - W'$ has the structure of a row-permutation of $W_{ICA}$. The structures can be generated by analyzing what happens when we permute the rows of $W'$. Remember that edges in $B'$ (and thus $W'$) are read column-to-row. Thus, row-permutations of $W'$ change the positions of the arrow-heads (targets), but not the arrow-tails (sources). Richardson proved that the operation of reversing a cycle preserves the set of entailed zero partial correlations, but did not consider distribution equivalence [11].

### 5.3 Adding the assumption of stability

In dynamical systems, "stable" models are ones in which the effects of one-time noise dissipate. For example, a model that has a single cycle whose cycle-product (product of coefficients of edges in the cycle) is $\geq 1$ is unstable, while one that has a single cycle whose cycle-product is between -1 and 1 is stable. On the other hand, if a positive feedback loop of cycle-product 2 is counteracted by a negative loop with cycle-product $-1.5$, then the model is stable, because the effective cycle-product is 0.5.

A general way to express stability is $\lim_{t \to \infty} B^t = 0$, which is mathematically equivalent to: for all eigenvalues $e$ of $B$, $|e| < 1$, in which $|z|$ means the modulus of $z$. This eigenvalues criterion is easy to compute.

Given only the coefficients between different variables, it is impossible to measure the stability of a SEM without assuming something about the self-loops. Therefore, in this section, it is assumed that the true model has no self-loops.

It is often the case that many of the SEMs output by LiNG-D are unstable. Since in many situations, the variables are assumed to be in equilibrium, we are often allowed to rule out unstable models.

In the remainder of this section, we will prove that if the SEM generating the population distribution has a graph in which the cycles are disjoint, then among the candidate SEMs output by LiNG-D, at most one will be stable.

**Theorem 3** *SEMs in the form of a simple cycle with a cycle-product $\pi$ such that $|\pi| \geq 1$ are unstable.*

**Proof:** Let $k$ be the length of the cycle. Then $B^k = \pi I$. Then for all integers $i$, $B^{ik} = \pi^i I$. So if $|\pi| \geq 1$, the entries of $B^{ik}$ do not get smaller than the entries of $B$ as $i$ increases. Thus, $B^t$ will not converge to 0 as $t \to \infty$. □

**Corollary 1:** *For SEMs in the form of a simple cycle, having a cycle-product $\geq 1$ is equivalent to having an eigenvalue $\geq 1$ (in modulus), which is equivalent to being unstable.*

**Theorem 4** *Suppose that there is a SEM $M$ with disjoint cycles with coefficient matrix $B$ and graph $G$ that entails a distribution $Q$, and a SEM $M_0 \neq M$ with graph $G_0$, coefficient matrix $B_0$, which is an admissible permutation of $M$ and also entails $Q$. Then $G_0$* also contains disjoint cycles, at least one of which is a reversal of a cycle $C$ in $G$, whose cycle-product is the inverse of the cycle-product of $C$.

**Proof:** Due to space limitations, the proof is just sketched here. Every permutation can be represented as a product of disjoint cyclic subpermutations of the form $a \rightarrow b \rightarrow \ldots m \rightarrow n \rightarrow a$, where $a \rightarrow b$ means $a$ gets mapped onto $b$. (Some cyclic subpermutations may be trivial, i.e. contain a single object mapped onto itself). Hence it suffices to prove the theorem for a single admissible cyclic row permutation of $B$. It can be shown that if a cyclic row permutation of $B$, $a \rightarrow b \rightarrow \ldots m \rightarrow n \rightarrow a$ is admissible, then $G$ contains the cycle $C$ equal to $a \leftarrow b \leftarrow \ldots m \leftarrow n \leftarrow a$, and $G_0$ contains the reversed cycle $C$ equal to $a \rightarrow b \rightarrow \ldots m \rightarrow n \rightarrow a$. Moreover, if $G_0$ contains two cycles that touch, so does $G$.

Consider $B_C$, the submatrix of $B$ that contains the coefficients of the edges in cycle $C$.

$$B_C = \begin{bmatrix} 0 & \ldots & 0 & b_{k,1} \\ b_{1,2} & 0 & \ldots & 0 \\ 0 & b_{2,3} & \ddots & 0 \\ 0 & 0 & \ddots & 0 \end{bmatrix}$$

Note that the cycle-product $\pi_C = b_{k,1} \prod_{i=0}^{k-1} b_{i,i+1}$.

$W_C = I - B_C$.

The "reversal" is the row-permutation in which the first row gets "rotated" into the bottom:

$$RowPermute(W_C) = \begin{bmatrix} -b_{1,2} & 1 & \ldots & 0 \\ 0 & -b_{2,3} & \ddots & 0 \\ 0 & 0 & \ddots & 1 \\ 1 & 0 & \ldots & -b_{k,1} \end{bmatrix}$$

Normalizing the diagonal to be all 1s, we get $W_{C'}$. Computing $B_{C'} = I - W_{C'}$, one can see that the cycle-product $\pi_{C'} = \frac{1}{b_{k,1}} \prod_{i=0}^{k-1} \frac{1}{b_{i,i+1}} = 1/\pi_C$. □

We will now show that for SEMs in which the cycles are disjoint, their stability only depends on the stability of the cycles.

**Theorem 5** *A SEM in which the cycles are disjoint is stable if and only if it has no unstable cycles.*

**Proof:** Let be $G$ be a SEM whose cycles are disjoint. Then $B_G$ can be written as a block-triangular matrix where each diagonal block is a cycle. The set of eigenvalues of a block-triangular matrix is the union of the sets of eigenvalues of the blocks in the diagonal (in this case, the eigenvalues of the cycles). Suppose a cycle of $G$ is unstable. Then it has an eigenvalue $\geq 1$ (in modulus). But since this is also an eigenvalue of $B_G$, it follows that $G$ is unstable. The other direction goes similarly. □

**Theorem 6** *If the true SEM is stable and has a graph in which the cycles are disjoint, then no other SEMs in the output of LiNG-D will be stable.*

**Proof:** Suppose the true SEM is stable and has a graph in which the cycles are disjoint. Call it $G$. Since, by Theorem 2, the output of LiNG-D are the admissible distribution-entailment equivalent alternatives to the true SEM, it suffices to show that all other admissible candidates are unstable.

By Theorem 5, all cycles in $G$ are stable. Let $H$ be an admissible alternative to $G$, such that $H \neq G$. By Theorem 4, $H$ will have at least one cycle $C$ reversed relative to $G$ and this reversed cycle will have a cycle product that is the inverse of the cycle product of $C$. By Corollary 1, the reversed matrix is not stable. Thus, by Theorem 5, $H$ is unstable.

Therefore, the only stable admissible alternative to $G$ is $G$ itself. □

It follows that if the true model's cycles are disjoint, then under the assumption that the true model is stable, we can fully identify it using a LiNG discovery algorithm (at most one SEM in the output of the LiNG discovery algorithm will be stable).

For example, consider the two candidate models shown in Fig. 3. By assuming that the true model is stable, one would select candidate #2. Since our simulation used a stable model, this is indeed the correct answer (see Fig. 1).

In general, however, there may be multiple stable models, and one cannot reliably select the correct one. When the cycles are not disjoint, it is easy to find examples for which there are multiple stable candidates.

The condition of disjoint cycles is sufficient, but not necessary: it is easy to come up with SEMs where we have exactly one stable SEM in the distribution-entailment equivalence class, despite intersecting cycles.

## 6 Discussion

We have presented Shimizu's approach for discovering LiNGAM SEMs, and generalized it to a method that discovers general LiNG SEMs. This improves upon the state-of-the-art on cyclic linear SEM discovery by outputting only the distribution-entailment equivalence class of SEMs, instead of the entire d-separation equiv-alence class; and by relaxing the faithfulness assumption. We have also shown that stability can be a powerful constraint, sometimes narrowing the candidates to a single SEM.

There are a number of questions that remain open for future research:

- The LiNG-D algorithm generates all admissible permutations. The worst-case time-complexity of n-Rooks is high, but can we do better than depth-first search for random instances? Is there an algorithm to efficiently search for the stable models, without going through all candidates? In the case where the cycles are disjoint, it is possible to just find the correct permutation for each cycle independently, but no such trick is known in general.

- How can prior information be incorporated into the algorithm?

- How can the algorithm be modified to allow the assumption of causal sufficiency to be relaxed? For the acyclic case, see [7].

- How can the algorithm be modified to allow for mixtures of non-Gaussian and Gaussian (or almost Gaussian) error terms? Hoyer et al [6] address this problem for the acyclic case.

- How could we integrate this method into mainstream dynamical systems research? Can the algorithm handle noisy dynamics and noisy observations? Could it be made to handle non-linear dynamics? What about self-loops of coefficient 1? How could one integrate this with methods that use non-equilibrium time-series data?

**Acknowledgements**

The authors wish to thank Anupam Gupta, Michael Dinitz and Cosma Shalizi. GL was partially supported by NSF Award No. REC-0537198.